\documentclass[
]{ceurart}

\usepackage{graphicx}
\usepackage{float}
\usepackage{multirow}

\begin{document}

\copyrightyear{2020}
\copyrightclause{Copyright for this paper by its authors.
  Use permitted under Creative Commons License Attribution 4.0
  International (CC BY 4.0).}

\conference{FIRE 2020: Forum for Information Retrieval Evaluation, December 16--20, 2020, Hyderabad, India}

\title{IIITG-ADBU@HASOC-Dravidian-CodeMix-FIRE2020: Offensive Content Detection in Code-Mixed Dravidian Text}

\author[1]{Arup Baruah}[%
email=arup.baruah@gmail.com
]
\address[1]{Indian Institute of Information Technology, Guwahati, India}

\author[1]{Kaushik Amar Das}[%
email=kaushikamardas@gmail.com
]

\author[1]{Ferdous Ahmed Barbhuiya}[%
email=ferdous@iiitg.ac.in
]

\author[2]{Kuntal Dey}[%
email=kuntal.dey@accenture.com
]
\address[2]{Accenture Technology Labs, Bangalore, India}

\begin{abstract}
  This paper presents the results obtained by our SVM and XLM-RoBERTa based classifiers in the shared task ``Dravidian-CodeMix-HASOC 2020''. The SVM classifier trained using TF-IDF features of character and word n-grams performed the best on the code-mixed Malayalam text. It obtained a weighted F1 score of 0.95 (1st Rank) and 0.76 (3rd Rank) on the YouTube and Twitter dataset respectively. The XLM-RoBERTa based classifier performed the best on the code-mixed Tamil text. It obtained a weighted F1 score of 0.87 (3rd Rank) on the code-mixed Tamil Twitter dataset. 
\end{abstract}


\maketitle

\section{Introduction}

The use of offensive language in social media text has become a new social problem. Such language can have a negative psychological impact on the readers. It can have adverse effect on the emotion and behavior of people. Hate speech has fueled riots in many places around the world. As such, it is important to keep social media free from offensive language. Considerable research has been performed on automated techniques for detecting offensive language. Among the many challenges that such systems have to tackle, the use of code-mixed text is another. Code-mixing is the phenomena of mixing words from more than one language in the same sentence or between sentences. 

The shared task ``Dravidian-CodeMix-HASOC 2020'' \cite{hasocdravidian--acm, hasocdravidian-ceur} is an attempt to promote research on offensive language detection in code-mixed text. This shared task is held as a sub-track of ``Hate Speech and Offensive Content Identification in Indo-European Languages (HASOC)'' at FIRE-2020. The shared task had two tasks. Task 1 required detection of offensive language in code-mixed Malayalam-English text from YouTube. Task 2 required detection of offensive language in code-mixed Tamil-English and Malayalam-English tweets. Both the tasks were binary classification problem where it was required to determine if the given text is offensive or not.

We participated in both the tasks. We used SVM and XLM-RoBERTa classifiers in our study. The SVM classifier was trained using TF-IDF features of character n-grams, word n-grams, and character and word n-grams combined. 

\section{Related Work}
Offensive language detection in English has witnessed the use of SVM \cite{malmasi, malmasi-2, magu, davidson, samghabadi}, Logistic Regression \cite{wulczyn, waseem-3, djuric, davidson,risch}, and deep learning techniques \cite{badjatiya, gamback, park,pavlopoulos-a, mehdad, baruah-1}. The main focus of \cite{magu} was to tackle the use of code words for obfuscating the hate words. Traditional machine learning and deep learning techniques have also been used in the detection of offensive language in code-mixed Hindi-English text \cite{santosh, bohra, kamble, sreelakshmi, mathur, baruah-2, baruah-3}. Work performed on code-mixed Tamil-English and Malayalam-English text includes corpus created for sentiment analysis for these two languages \cite{bharathi-t, bharathi-m}. \cite{bharathi-th} focused on machine translation of code-mixed text in Dravidian languages. It was found that removal of code-mixing improves the quality of machine translation.

\section{Dataset}

Table \ref{tab:dataset-statistics} shows the statistics of the dataset provided as part of this shared task. The instances in the dataset were labeled as ``not offensive'' (NOT) or ``offensive'' (OFF). Task 1 was conducted for Malayalam language only. The source of the dataset for this task was YouTube. As can be seen from the table, this dataset is imbalanced with about 83\% labeled as NOT. Task 2 was conducted for both Tamil and Malayalam languages. The source of the datasets for this task was Twitter. As can be seen from the tables, the dataset for this task was balanced. Train, development, and test set was provided for Task 1. For task 2, only train and test set was provided. We created the development set for Task 2, by doing a stratified split and retaining 85\% of the dataset for training and 15\% as development dataset.

\begin{table}
	\caption{Data set statistics}
	\label{tab:dataset-statistics}
	\begin{center}
		\begin{tabular}{|c c c c c c c c|}
		    \hline
			\textbf{Label} & \multicolumn{3}{c}{\textbf{Task 1 - Malayalam}} & \multicolumn{2}{c}{\textbf{Task 2 - Tamil}} & \multicolumn{2}{c|}{\textbf{Task 2 - Malayalam}} \\
			 & \textbf{Train} & \textbf{Dev} & \textbf{Test} & \textbf{Train} & \textbf{Test} & \textbf{Train} & \textbf{Test} \\
			\hline 
			\hline
			NOT & 2633 & 328 & 334 & 2020 & 465 & 2047 & 488 \\
			& (82.3\%) & (82\%) & (83.5\%) & (50.5\%) & (49.5\%) & (51.2\%) & (48.8\%) \\
			OFF & 567 & 72 & 66 & 1980 & 475 & 1953 & 512 \\
			& (17.7\%) & (18\%) & (16.5\%) & (49.5\%) & (50.5\%) & (48.4\%) & (51.2\%) \\
			Total & 3200 & 400 & 400 & 4000 & 940 & 4000 & 1000 \\
			\hline
		\end{tabular}
	\end{center}
\end{table}

\section{Methodology}
In this study we used SVM and XLM-RoBERTa based classifiers. The SVM classifier was trained using TF-IDF features of character n-grams, word n-grams, and combination of character and word n-grams. In our study, we used character n-grams of size 1 to 6, and word n-grams of size 1 to 3.

XLM-RoBERTa model \cite{xlmroberta} is based on the RoBERTa model \cite{roberta}. RoBERTa model is based on the transformer architecture.  XLM-RoBERTa is a multi-lingual model trained on 100 different languages including Tamil and Malayalam. In our study, we used the pre-trained base model. The Adam optimizer with weight decay was used during training. The learning rate and epsilon parameter for the optimizer were set to 2e-5 and 1e-8 respectively. We used the class provided by HuggingFace Transformers library \footnote{~\url{https://huggingface.co/transformers/}} for sequence classification in our study. This class provides a linear layer on top of the pooled output to perform the binary classification.

\section{Results}

\begin{table*}
\caption{Dev Set Results}
\label{tab:dev-set-results}
\center
\begin{tabular}{|lllll|}
\hline
\bf Task & \bf System & \bf Precision & \bf Recall & \bf F1 \\ 
& & \bf (Weighted) & \bf (Weighted) & \bf (Weighted) \\
\hline
\hline
Task 1 Malayalam & SVM (char) & 0.9187	& 0.9175 & 0.9096 \\
Task 1 Malayalam & SVM (word) & 0.9138	& 0.9075 & 0.8950 \\
Task 1 Malayalam & SVM (char + word) & \textbf{0.9330}	& \textbf{0.9325} & 0.9278 \\
Task 1 Malayalam & XLM-RoBERTa & 0.9305	& \textbf{0.9325} & \textbf{0.9307} \\
\hline
Task 2 Tamil & SVM (char) & 0.8650	& 0.8633 & 0.8630 \\
Task 2 Tamil & SVM (word) & \textbf{0.8733}	& \textbf{0.8717} & \textbf{0.8714} \\
Task 2 Tamil & SVM (char + word) & 0.8617 & 0.8600 & 0.8597 \\
Task 2 Tamil & XLM-RoBERTa & 0.8651	& 0.8650 & 0.8650 \\
\hline
Task 2 Malayalam & SVM (char) & 0.7519	& 0.7500 & 0.7490 \\
Task 2 Malayalam & SVM (word) & 0.7190	& 0.7100 & 0.7056 \\
Task 2 Malayalam & SVM (char + word) & \textbf{0.7630}	& \textbf{0.7617} & \textbf{0.7610} \\
Task 2 Malayalam & XLM-RoBERTa & 0.5732	& 0.5483 & 0.5171 \\
\hline
\end{tabular}
\end{table*}

\begin{table*}
\caption{Test Set Results}
\label{tab:test-set-results}
\center
\begin{tabular}{|llllll|}
\hline
\bf Task & \bf System & \bf Precision & \bf Recall & \bf F1 & \bf Rank \\ 
& & \bf (Weighted) & \bf (Weighted) & \bf (Weighted) & \\
\hline
\hline
Task 1 Malayalam & SVM (char + word) & \textbf{0.9505}	& \textbf{0.9500} & \textbf{0.9471} & 1st \\
Task 1 Malayalam & XLM-RoBERTa & 0.9241	& 0.9250 & 0.9245 & - \\
\hline
Task 2 Tamil & SVM (word) & 0.8524	& 0.8521 & 0.8520 & - \\
Task 2 Tamil & XLM-RoBERTa & \textbf{0.8680}	& \textbf{0.8670} & \textbf{0.8669} & 3rd \\
\hline
Task 2 Malayalam & SVM (char + word) & \textbf{0.7686}	& \textbf{0.7630} & \textbf{0.7623} & 3rd \\
Task 2 Malayalam & XLM-RoBERTa & 0.6181	& 0.5800 & 0.5337 & -\\
\hline
\end{tabular}
\end{table*}

\begin{table*}
	\caption{Confusion Matrices of the submitted classifiers on the Test Set}
	\begin{center}
		\begin{tabular}{ | c  c  c  c  c  c  c  c  c  c  c  c  c |}			
			\hline
			
			& \multicolumn{4}{c}{\textbf{Task 1 (Malayalam)}} & \multicolumn{4}{c}{\textbf{Task 2 (Tamil)}} & 
			\multicolumn{4}{c|}{\textbf{Task 2 (Malayalam)}} \\			
			& \multicolumn{2}{c}{\textbf{SVM}} & \multicolumn{2}{c}{\textbf{XLM-}} & \multicolumn{2}{c}{\textbf{SVM}} &
			\multicolumn{2}{c }{\textbf{XLM-}} & 
			\multicolumn{2}{c}{\textbf{SVM}} &
			\multicolumn{2}{c|}{\textbf{XLM-}} \\

			& \multicolumn{2}{c}{\textbf{char+word}} & \multicolumn{2}{c}{\textbf{RoBERTa}} & \multicolumn{2}{c}{\textbf{word}} &
			\multicolumn{2}{c}{\textbf{RoBERTa}} & 
			\multicolumn{2}{c}{\textbf{char+word}} &
			\multicolumn{2}{c|}{\textbf{RoBERTa}} \\
			
			& \multicolumn{2}{c}{\textbf{pred}} & \multicolumn{2}{c}{\textbf{pred}} & \multicolumn{2}{c}{\textbf{pred}} &
			\multicolumn{2}{c}{\textbf{pred}} & 
			\multicolumn{2}{c}{\textbf{pred}} &
			\multicolumn{2}{c|}{\textbf{pred}} \\
			\hline
			\hline
			
			& \textbf{NOT} & \textbf{OFF} & \textbf{NOT} & \textbf{OFF} & \textbf{NOT} & \textbf{OFF} & \textbf{NOT} & \textbf{OFF} & \textbf{NOT} & \textbf{OFF} & \textbf{NOT} & \textbf{OFF} \\
			\textbf{NOT} & 332 & 2 & 320 & 14 & 389 & 76 & 390 & 75 & 403 & 85 & 127 & 361 \\
			\textbf{OFF} & 18 & 48 & 16 & 50 & 63 & 412 & 50 & 425 & 152 & 360 & 59 & 453 \\ \hline
		\end{tabular}
		\label{tab:confusion-test-set}	
	\end{center}
\end{table*}

Table \ref{tab:dev-set-results} shows the results obtained by our SVM and XLM-RoBERTa classifiers on the development set. For task 1, the development set was provided as part of the dataset. For task 2, the development set was created by performing a stratified split on the train set. 15\% of the train set was set aside as the development set. The XLM-RoBERTa classifier performed the best with a weighted F1 score of 0.9307 in the development set for task 1. Among the SVM classifiers, the one trained using the combination of TF-IDF features of character and word n-grams performed the best with a weighted F1 score of 0.9278. 

In task 2 dev set, the SVM classifier trained using the TF-IDF features of word n-grams performed the best for code-mixed Tamil-English text. It obtained a weighted F1 score of 0.8714. The XLM-RoBERTa classifier obtained a weighted F1 score of 0.8650 and was the second best performing classifier on the dev set for this task. For code-mixed Malayalam-English text of the task 2 dev set, the best performing classifier was the SVM classifier trained using the combination of TF-IDF features of character and word n-grams. It obtained a weighted F1 score of 0.7610. The XLM-RoBERTa classifier obtained a weighted F1 score of 0.5171 and was the worst performing classifier for this task.

Table \ref{tab:test-set-results} shows the results that our submitted classifiers obtained on the test set. The SVM classifiers mentioned in this table are the only one submitted for the tasks. These classifiers were selected based on their performance on the development set. As can be seen from the table, the SVM classifier trained on the combination of TF-IDF features of character and word n-grams performed the best in task 1 with as weighted F1 score of 0.9471. It obtained the 1st rank for the task. XLM-RoBERTa was the best performing classifier for the Tamil-English dataset of task 2. It was a weighted F1 score of 0.8669 and obtained the 3rd rank for the task. The SVM classifier trained on the combination of TF-IDF features of character and word n-grams again performed the best for the Malayalam-English dataset of task 2 with a weighted F1 score of 0.7623. It obtained the 3rd rank for the task.  Table \ref{tab:confusion-test-set} shows the confusion matrices obtained on the test set by classifiers submitted for the shared task.

\section{Conclusion}
We used the SVM and XLM-RoBERTa based classifiers to detect offensive language in code-mixed Tamil-English and Malayalam-English text. In our study, the SVM classifier trained using combination of TF-IDF features of character and word n-grams performed the best for code-mixed Malayalam-English text (both YouTube and Twitter dataset). This classifier obtained the weighted F1 score of 0.95 (1st rank) and 0.76 (3rd rank) for Task 1 and Task 2 (Malayalam) respectively. The XLM-RoBERTa based classifier performed the best for the Tamil-English dataset of Task 2 and obtained an weighted F1 score of 0.87 (3rd rank) for the task. On comparing the performance of our SVM models on the YouTube and Twitter data for Malayalam language, we can observe that the performance of the classifier degraded considerably for the Twitter dataset. Whether this degradation is due to the type of language used in Twitter conversation, length of the text etc. can be performed as a future study.

\begin{acknowledgments}
  Supported by Visvesvaraya PhD Scheme, MeitY, Govt. of India, MEITY-PHD-3050.
\end{acknowledgments}

\bibliography{sample-ceur}

\end{document}